%% file: main4.tex
\documentclass[10pt,twocolumn,letterpaper]{article}
\usepackage{cvpr}
\input{preamble}
\definecolor{cvprblue}{rgb}{0.21,0.49,0.74}
\usepackage[pagebackref,breaklinks,colorlinks,allcolors=cvprblue]{hyperref}
\usepackage{booktabs}
\usepackage{multirow}
\usepackage{graphicx}
\usepackage{xcolor}
\usepackage{amsmath,amssymb}
\usepackage{pifont}
\usepackage{tikz}
\usetikzlibrary{arrows.meta,positioning,fit,calc,shapes.geometric}
\newcommand{\cmark}{\ding{51}}
\newcommand{\xmark}{\ding{55}}

\def\paperID{***}
\def\confName{CVPR}
\def\confYear{2027}

\title{ScriptHOI: Learning Scripted State Transitions for Open-Vocabulary Human-Object Interaction Detection}

\author{Minh Anh Nguyen,
Quang Huy Tran,
Bao Ngoc Le, SuiYang Guang,
Tuan Kiet Pham,
Linh Chi Vo\\ Phenikaa University}

\begin{document}
\maketitle

\begin{abstract}
Open-vocabulary human-object interaction (HOI) detection requires recognizing interaction phrases that may not appear as annotated categories during training. Recent vision-language HOI detectors improve semantic transfer by matching human-object features with text embeddings, but their predictions are often dominated by object affordance and phrase-level co-occurrence. As a result, a model may predict \textit{cut cake} from the presence of a knife and a cake without verifying whether the hand, tool, target, contact pattern, and object state jointly support the action. We propose \textbf{ScriptHOI}, a structured framework that represents each interaction phrase as a soft scripted state transition. Rather than treating a phrase as a single class token, ScriptHOI decomposes it into body-role, contact, geometry, affordance, motion, and object-state slots. A visual state tokenizer parses each detected human-object pair into corresponding state tokens, and a slot-wise matcher estimates both script coverage and script conflict. These two quantities calibrate HOI logits, expose missing visual evidence, and provide training constraints for incomplete annotations. To avoid suppressing valid but unannotated interactions, we further introduce interval partial-label learning, which constrains unannotated candidates with script-derived lower and upper probability bounds instead of assigning closed-world negatives. A counterfactual script contrast loss swaps individual script slots to discourage object-only shortcuts. Experiments on HICO-DET, V-COCO, and open-vocabulary HOI splits show that ScriptHOI improves rare and unseen interaction recognition while substantially reducing affordance-conflict false positives.
\end{abstract}

\section{Introduction}

Human-object interaction detection localizes humans and objects and predicts the interaction that connects them~\cite{gupta2015visual,chao2018learning,gkioxari2018detecting,gao2018ican}. It is a key component of human-centric scene understanding, activity recognition, visual question answering, image retrieval, assistive robotics, embodied navigation, and image-to-graph reasoning~\cite{krishna2017visual,antol2015vqa,johnson2015image,xu2017scene,he2021exploiting,he2023toward,zakari2025vqa}. The difficulty of HOI detection lies in the fact that an interaction is not determined by the object category alone. The same object can participate in many visually distinct actions. A cup can be \textit{held}, \textit{drunk from}, \textit{washed}, \textit{filled}, or simply \textit{near} a person. A bicycle can be \textit{ridden}, \textit{pushed}, \textit{repaired}, or \textit{parked beside}. Correct recognition therefore requires reasoning about human pose, body-part involvement, contact, spatial layout, object affordance, surrounding context, and the visible state of the object.

Early HOI detectors classify human-object pairs using union-region appearance, pairwise geometry, pose cues, attention, language priors, and graph reasoning~\cite{fang2018pairwise,qi2018learning,li2019transferable,ulutan2020vsgnet,hou2020visual,gao2020drg}. Recent methods adopt one-stage detection, transformer decoders, query-based pair reasoning, set prediction, compositional verb-object modeling, and vision-language pretraining~\cite{tamura2021qpic,kim2021hotr,zhou2021end,zhang2021mstr,liao2022cdn,zhang2022gen,park2022consistency}. For open-vocabulary HOI detection, CLIP-style representations, prompt learning, locality-aware queries, and universal interaction embeddings enable transfer to unseen verbs and unseen triplets~\cite{radford2021learning,jia2021scaling,li2022blip,mao2023clip4hoi,cao2023universal,yang2024towards}. Related adaptive prompt-learning studies in vision-language models further show that prompt structure can improve robustness when cross-modal evidence is incomplete~\cite{dai2024muap}.

Despite these advances, most open-vocabulary HOI detectors still use a flat phrase representation. A candidate interaction such as \textit{person cutting cake with knife} is encoded as one text label and compared against a global human-object feature. This design is attractive because it inherits semantic knowledge from large vision-language models, but it hides the visual requirements that make the interaction true. Cutting a cake involves a hand or tool, a target object, near-contact, an appropriate spatial configuration, and a plausible state change. A global compatibility score can learn that \textit{knife}, \textit{cake}, and \textit{cut} often co-occur, yet it has no explicit mechanism to check whether the person is using the knife or whether the knife is only present in the scene. This creates a common failure mode in open-vocabulary HOI: the model transfers object affordance but not interaction evidence.

We address this limitation by rethinking the representation of an interaction phrase. Our central observation is that many HOI categories can be expressed as compact \emph{scripted state transitions}. A script is not a hand-crafted symbolic program. Instead, it is a differentiable, slot-based description of the visual conditions that should hold for an interaction to be plausible. For example, \textit{hold cup} emphasizes hand involvement, direct contact, close geometry, graspable affordance, and a held object state. In contrast, \textit{look at television} emphasizes head or gaze orientation and does not require contact. This view exposes the latent structure of an interaction and provides a natural interface between language phrases and localized visual evidence.

We propose \textbf{ScriptHOI}, a scripted state-transition framework for open-vocabulary HOI detection. Given a detected human-object pair, ScriptHOI parses the pair into visual state tokens covering body role, contact, geometry, affordance, motion proxy, object state, and local context. Given an interaction phrase, a script parser converts the phrase into soft slot distributions. A script-state matcher then aligns the phrase script with the visual state tokens and produces two complementary scores. The coverage score measures whether the visual pair satisfies the required script slots, while the conflict score measures whether important slots are contradicted. The final HOI score is obtained by calibrating the base detector logit with these two scores.

This structured representation also provides a principled way to learn from incomplete HOI annotations. Standard datasets such as HICO-DET and V-COCO are not exhaustively labeled. A human-object pair may have multiple valid interactions, but only a subset is annotated~\cite{chao2018learning,gupta2015visual}. Treating every unannotated interaction as negative incorrectly penalizes valid missing labels, while raw pseudo-labeling can reinforce spurious object co-occurrence. ScriptHOI introduces interval partial-label learning for this setting. Annotated interactions are trained as positives. Unannotated candidates receive lower and upper probability bounds derived from script coverage and script conflict. Thus, visually compatible missing interactions are not forced down, and visually contradicted interactions are prevented from becoming overconfident.

Our contributions are summarized as follows:
\begin{itemize}
    \item We introduce scripted state-transition tokens for open-vocabulary HOI detection, replacing flat phrase matching with structured script-to-state alignment.
    \item We design a visual state tokenizer that decomposes each human-object pair into body-role, contact, geometry, affordance, motion-proxy, object-state, and context tokens.
    \item We propose a slot-wise script-state matcher that estimates both coverage and conflict, enabling evidence-aware calibration and reducing affordance-only shortcuts.
    \item We develop interval partial-label learning for incomplete HOI annotations, where unannotated interactions are constrained by script-derived probability intervals rather than treated as closed-world negatives.
    \item We introduce counterfactual script contrast and an affordance-conflict evaluation protocol to measure whether models recognize interaction evidence rather than object affordance alone.
\end{itemize}

\section{Related Work}

\subsection{Human-Object Interaction Detection}
HOI detection has been studied with two-stage pair classification, interactiveness estimation, pose-guided interaction modeling, visual-semantic graphs, and one-stage detectors~\cite{gupta2015visual,chao2018learning,gkioxari2018detecting,gao2018ican,fang2018pairwise,qi2018learning,li2019transferable,ulutan2020vsgnet,hou2020visual,gao2020drg,zhong2021glance}. Transformer-based methods further formulate HOI detection as query-based set prediction, direct triplet decoding, or end-to-end pair reasoning~\cite{carion2020end,tamura2021qpic,kim2021hotr,zhou2021end,zhang2021mstr,liao2022cdn,zhang2022upt,zhang2022gen,park2022consistency}. Recent studies explore hierarchical tuple correlations and unified scene-graph and HOI reasoning~\cite{he2021exploiting,he2023toward,hu2026exploring}. The success of cross-view interactive transformer designs also suggests that explicit information exchange between paired visual streams can strengthen fine-grained representation learning~\cite{zhang2024cviformer}. ScriptHOI is complementary to these detectors. It does not replace the localization or pair decoding backbone, but augments the interaction classification branch with structured script evidence.

\subsection{Open-Vocabulary Vision-Language Learning}
Large-scale vision-language pretraining provides transferable representations for recognition, detection, grounding, and segmentation~\cite{radford2021learning,jia2021scaling,yao2021filip,li2022blip,alayrac2022flamingo,yu2022coca,li2023blip2,gu2021open,zhong2022regionclip,li2022grounded,liu2023grounding,kirillov2023segment}. Prompt learning and region-language alignment further adapt these models to dense visual understanding~\cite{zhou2022coop,zhou2022cocoop}. In HOI detection, CLIP-based phrase embeddings and locality-aware interaction queries improve zero-shot and open-vocabulary generalization~\cite{zhang2022gen,mao2023clip4hoi,cao2023universal,yang2024towards}. These methods usually align a visual pair with a whole phrase. ScriptHOI instead translates the phrase into state-transition slots and verifies each slot against localized visual tokens.

\subsection{Structured Representation and Transfer}
Structured representation learning has also shown that compact and transferable embeddings are useful when supervision is limited or domains change. Semi-supervised attributed network embedding with attention-based quantisation and differentiable deep quantization learn discriminative discrete codes under partial supervision~\cite{he2020sneq,he2021semisupervised}, while transferable discrete network embedding with hierarchical knowledge distillation improves multi-domain generalization~\cite{he2023transferable}. Although these works address graph and representation learning rather than HOI detection, they motivate a similar principle: a structured latent space can reduce shortcut learning and improve transfer. ScriptHOI follows this principle by decomposing each human-object pair into script-aligned state tokens instead of relying on a single flat pair representation.

\subsection{Compositional and State-Aware Reasoning}
Compositional recognition studies how objects, attributes, relations, and actions generalize beyond observed combinations~\cite{misra2017red,naeem2021learning,purushwalkam2020task,li2023compositional}. Scene graph generation and relation reasoning also use semantic composition, prompt-based open-vocabulary transfer, state-aware prediction, lifelong in-context learning, and spatial-aware denoising to improve visual relationship understanding~\cite{krishna2017visual,zellers2018motifs,tang2020unbiased,he2020learning,he2021semantic,he2022state,he2022towards,he2024towardslifelong,hu2025spade}. Causal and counterfactual learning reduce reliance on dataset bias by enforcing invariance under interventions~\cite{pearl2009causality,arjovsky2019invariant,tang2020unbiased,wang2020vcl,yin2025knowledge}. ScriptHOI follows this direction but targets the open-vocabulary HOI setting. It uses counterfactual script contrast to change one interaction slot at a time, forcing the detector to respond to body role, contact, geometry, affordance, and state evidence rather than object co-occurrence.

\subsection{Learning with Incomplete Supervision}
Sparse labels and missing annotations are common in visual relationship and HOI datasets. Positive-unlabeled learning and partial-label learning address the case where an unobserved label cannot be safely treated as negative~\cite{elkan2008learning,kiryo2017positive,feng2020provably,lv2020progressive}. Related multimodal studies handle missing information through prompt distillation, dynamic fusion, incomplete-modality learning, rationale generation, robust prompt tuning, and adaptive prompt learning~\cite{dai2026anchor,wei2026unbiased,dai2025unbiasedmissing,dong2025unbiased,dai2025robustpt,dai2024muap,daiTowardsIncomplete}. Recent dataset-distillation work further indicates that dynamic retrieval and persistent structural anchors can improve transfer when the available supervision is sparse or biased~\cite{li2026fixed}. ScriptHOI adapts these ideas to HOI detection by deriving probability intervals from structured script evidence. The interval is high only when the visual pair covers the script and low only when important script requirements are violated.

\section{Method}

\subsection{Overview}

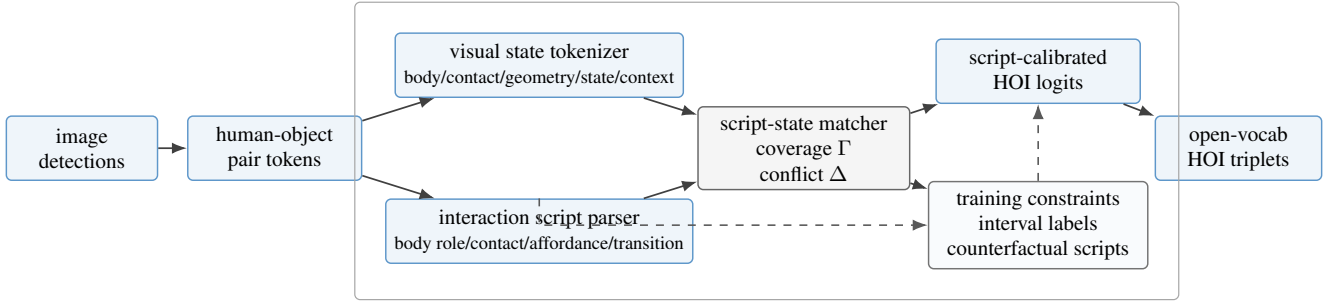
\begin{figure*}[t]
\centering
\begin{tikzpicture}[
    font=\footnotesize,
    stage/.style={draw=black!70, rounded corners=2pt, fill=black!2, line width=0.55pt, align=center, minimum height=8.5mm},
    main/.style={stage, fill=cvprblue!8, draw=cvprblue!80},
    aux/.style={stage, fill=black!4, draw=black!60},
    train/.style={stage, fill=cvprblue!3, draw=black!55},
    arrow/.style={-{Latex[length=2.0mm]}, line width=0.65pt, draw=black!75},
    dashedarrow/.style={-{Latex[length=2.0mm]}, dashed, line width=0.6pt, draw=black!65}
]

\node[main, minimum width=20mm] (img) at (-7.2,0) {image\\detections};
\node[main, minimum width=23mm] (pair) at (-4.65,0) {human-object\\pair tokens};
\node[main, minimum width=34mm] (state) at (-1.15,1.10) {visual state tokenizer\\{\scriptsize body/contact/geometry/state/context}};
\node[main, minimum width=34mm] (script) at (-1.15,-1.10) {interaction script parser\\{\scriptsize body role/contact/affordance/transition}};
\node[aux, minimum width=28mm] (match) at (2.35,0) {script-state matcher\\coverage $\Gamma$\\conflict $\Delta$};
\node[train, minimum width=29mm] (learn) at (5.45,-1.02) {training constraints\\interval labels\\counterfactual scripts};
\node[main, minimum width=27mm] (score) at (5.45,1.02) {script-calibrated\\HOI logits};
\node[main, minimum width=22mm] (out) at (8.10,0) {open-vocab\\HOI triplets};

\draw[arrow] (img) -- (pair);
\draw[arrow] (pair) -- (state);
\draw[arrow] (pair) -- (script);
\draw[arrow] (state) -- (match);
\draw[arrow] (script) -- (match);
\draw[arrow] (match) -- (score);
\draw[arrow] (score) -- (out);
\draw[arrow] (match) -- (learn);
\draw[dashedarrow] (script) |- (learn);
\draw[dashedarrow] (learn) -- (score);

\node[draw=black!35, rounded corners=2pt, fit=(state)(script)(match)(learn)(score), inner sep=4mm, line width=0.45pt] {};
\end{tikzpicture}
\caption{Overall framework of ScriptHOI. The visual branch parses a detected human-object pair into state tokens, while the language branch parses each candidate phrase into a soft interaction script. Slot-wise matching produces a coverage score $\Gamma$ and a conflict score $\Delta$, which calibrate HOI logits and define interval constraints for unannotated interactions.}
\label{fig:framework}
\end{figure*}

Most open-vocabulary HOI detectors compare a human-object representation with a phrase embedding. This provides semantic transfer but does not reveal which visual conditions are responsible for a prediction. ScriptHOI instead represents an interaction as a short state-transition script and verifies whether the pair satisfies the script. As shown in Fig.~\ref{fig:framework}, the framework has three components: a visual state tokenizer, an interaction script parser, and a script-state matcher. The matcher outputs script coverage and script conflict, which are used for both inference calibration and training with incomplete labels.

\subsection{Problem Formulation}

Let $\mathcal{I}$ be an image. A detector returns human instances $\mathcal{H}=\{h_i\}_{i=1}^{N_h}$ and object instances $\mathcal{O}=\{o_j\}_{j=1}^{N_o}$. Each human instance has a box $b_i^h$, feature $f_i^h$, and pose tokens $p_i$. Each object instance has category $c_j$, box $b_j^o$, feature $f_j^o$, and optional mask $m_j^o$. For each pair $(h_i,o_j)$, the model predicts interactions from an open phrase vocabulary $\mathcal{V}^{open}$. The observed label $y_{ij}^{v}$ is sparse. If $y_{ij}^{v}=1$, the interaction $v$ is annotated as present. If $y_{ij}^{v}=0$, the interaction may be a true negative or a missing positive. We therefore estimate
\begin{equation}
    p_{\theta}(z_{ij}^{v}=1\mid \mathcal{I},h_i,o_j,v),
\end{equation}
where $z_{ij}^{v}$ denotes the latent interaction state. Unlike closed-world training, we do not assume that every unobserved interaction is negative.

A base HOI decoder produces a pair logit $s_{ij}^{v}$. ScriptHOI augments this logit using two script quantities: a coverage score $\Gamma_{ij}^{v}$ that measures whether the visual tokens satisfy the script, and a conflict score $\Delta_{ij}^{v}$ that measures whether required script slots are violated.

\subsection{Soft Interaction Scripts}

For each phrase $v$, ScriptHOI constructs a soft script
\begin{equation}
    \pi_v=\{\pi_v^{body},\pi_v^{contact},\pi_v^{geom},
    \pi_v^{aff},\pi_v^{motion},\pi_v^{state}\}.
\end{equation}
The six slots describe body-part role, contact mode, spatial geometry, object affordance, motion tendency, and expected object-state transition. These slots are chosen because they correspond to visual conditions that often distinguish interactions sharing the same object. For example, \textit{hold cup} and \textit{drink from cup} both involve a cup and hand contact, but the latter additionally requires a mouth-related body role and a different relative geometry. Likewise, \textit{ride bicycle}, \textit{push bicycle}, and \textit{repair bicycle} share the object but require different body configurations and contact patterns.

Scripts are initialized from dataset verbs, caption-mined interaction phrases, and a compact vocabulary of slot names. They are not used as hard rules. Given a phrase embedding $t_v$, each slot is represented as a distribution over possible slot values:
\begin{equation}
    P_{\pi}^{k}(\cdot|v)=\mathrm{Softmax}(W_{\pi}^{k}t_v),
    \qquad k\in\mathcal{K},
\end{equation}
where $\mathcal{K}$ denotes the set of script slots. This soft design is important for open vocabulary recognition. A specific phrase such as \textit{eat apple} has a peaked script, while a broad phrase such as \textit{use object} can admit multiple body parts, contact types, and state outcomes.

\subsection{Visual State Tokenization}

The visual branch converts each human-object pair into state tokens aligned with the script slots. We first construct a pair sequence
\begin{equation}
    X_{ij}=[f_i^h,f_j^o,f_{ij}^{u},p_i,g_{ij},r_{ij}^{part},r_{ij}^{ctx}],
\end{equation}
where $f_{ij}^{u}$ is the union-region feature, $g_{ij}$ is relative geometry, $r_{ij}^{part}$ denotes body-part features around the hands, feet, head, and torso, and $r_{ij}^{ctx}$ contains local scene-context tokens. A lightweight transformer with learnable slot queries outputs
\begin{equation}
    \mathcal{S}_{ij}=\{S_{ij}^{body},S_{ij}^{contact},S_{ij}^{geom},
    S_{ij}^{aff},S_{ij}^{motion},S_{ij}^{state}\}.
\end{equation}

Each token is encouraged to specialize in one visual factor. The body token attends to pose and part crops. The contact token aggregates hand-object distance, body-part distance, mask overlap, and nearest-boundary distance. The geometry token captures relative position, scale, and containment. The affordance token uses object category and appearance to model what actions the object can support. The motion token captures static cues such as limb extension, grasp direction, and displacement hints. The state token captures visible object attributes such as open, closed, worn, held, cut, filled, or placed. By isolating affordance as one token rather than allowing it to dominate the full pair representation, ScriptHOI can separate object plausibility from actual interaction evidence.

\subsection{Script-State Matching}

Given a phrase script and a visual state, ScriptHOI computes slot-wise compatibility. For slot $k$, we define
\begin{equation}
    m_{ij}^{v,k}=
    \mathrm{sim}(W_s^k S_{ij}^{k}, W_{\pi}^k P_{\pi}^{k}(\cdot|v)),
\end{equation}
where $\mathrm{sim}(\cdot,\cdot)$ denotes normalized similarity after projection. Not all slots are equally important for every interaction. Contact is crucial for \textit{hold}, \textit{carry}, and \textit{cut}, but is weak for \textit{look at}. We therefore predict a phrase-dependent slot reliability
\begin{equation}
    \rho_v^k=\sigma(w_{\rho}^{k\top}t_v).
\end{equation}
The coverage score is a weighted geometric aggregation of slot compatibilities:
\begin{equation}
    \Gamma_{ij}^{v}
    =\exp\left(
    \frac{\sum_{k\in\mathcal{K}}\rho_v^k \log(\sigma(m_{ij}^{v,k})+\epsilon)}
    {\sum_{k\in\mathcal{K}}\rho_v^k+\epsilon}
    \right).
\end{equation}
The geometric form makes the score sensitive to missing required slots. A high value indicates that the visual pair jointly covers the important script requirements.

Coverage alone is insufficient because an object may afford an action even when the visual evidence contradicts the script. We thus compute a conflict score from weighted slot residuals:
\begin{equation}
    \Delta_{ij}^{v}=
    \sigma\left(w_{\Delta}^{\top}
    [\rho_v^k(1-\sigma(m_{ij}^{v,k}))]_{k\in\mathcal{K}}\right).
\end{equation}
For instance, if a phrase requires hand contact but all hand-object distances are large, the conflict score increases even when the object is semantically compatible with the verb. The final script-calibrated logit is
\begin{equation}
    \hat{s}_{ij}^{v}=s_{ij}^{v}
    +\lambda_{\Gamma}\log(\Gamma_{ij}^{v}+\epsilon)
    -\lambda_{\Delta}\Delta_{ij}^{v}.
\end{equation}
This calibration rewards visually covered scripts and penalizes contradicted scripts.

\subsection{Learning with Incomplete HOI Labels}

ScriptHOI is trained with four loss terms. The first term is supervised HOI detection on annotated triplets:
\begin{equation}
    \mathcal{L}_{hoi}
    =\mathbb{E}_{y_{ij}^{v}=1}[-\log\sigma(\hat{s}_{ij}^{v})].
\end{equation}
When the detector is optimized end-to-end, standard human and object localization losses are inherited from the base HOI decoder.

The second term handles unannotated interactions through interval partial-label learning. For an unannotated candidate, strong script coverage suggests that the candidate should not be pushed below a lower bound, while strong conflict suggests that it should not exceed an upper bound. We define
\begin{equation}
    \ell_{ij}^{v}=\alpha_{\ell}\Gamma_{ij}^{v}(1-\Delta_{ij}^{v}),
    \qquad
    u_{ij}^{v}=1-\alpha_u\Delta_{ij}^{v}(1-\Gamma_{ij}^{v}).
\end{equation}
With $p_{ij}^{v}=\sigma(\hat{s}_{ij}^{v})$, the interval loss is
\begin{equation}
    \mathcal{L}_{ipl}
    =\mathbb{E}_{y_{ij}^{v}=0}
    [\max(0,\ell_{ij}^{v}-p_{ij}^{v})^2+
    \max(0,p_{ij}^{v}-u_{ij}^{v})^2].
\end{equation}
This loss leaves uncertain candidates weakly constrained. It also prevents two common errors: suppressing visually valid missing positives and promoting visually contradicted pseudo-positives.

The third term is counterfactual script contrast. For a positive or high-coverage candidate $v$, we construct counterfactual phrases $\mathcal{C}(v)$ by changing one script slot while preserving other slots when possible. Examples include replacing \textit{hold cup} with \textit{kick cup}, or replacing \textit{wear hat} with \textit{hold hat} when the hat is located on the head. The loss is
\begin{equation}
    \mathcal{L}_{csc}
    =-\log
    \frac{\exp(\hat{s}_{ij}^{v}/\tau)}
    {\exp(\hat{s}_{ij}^{v}/\tau)+
    \sum_{\tilde{v}\in\mathcal{C}(v)}\exp(\hat{s}_{ij}^{\tilde{v}}/\tau)}.
\end{equation}
Since the counterfactual phrase often shares the same object, this loss discourages shortcuts based on object category or phrase co-occurrence.

The fourth term regularizes script-state alignment on annotated positives:
\begin{equation}
    \mathcal{L}_{align}
    =\mathbb{E}_{y_{ij}^{v}=1}
    \left[-\log(\Gamma_{ij}^{v}+\epsilon)
    -\log(1-\Delta_{ij}^{v}+\epsilon)\right].
\end{equation}
It encourages annotated positives to be both well covered and low in conflict. The complete objective is
\begin{equation}
    \mathcal{L}
    =\mathcal{L}_{hoi}
    +\lambda_{ipl}\mathcal{L}_{ipl}
    +\lambda_{csc}\mathcal{L}_{csc}
    +\lambda_{align}\mathcal{L}_{align}.
\end{equation}

\subsection{Inference and Script-Aware Suppression}

At inference, ScriptHOI ranks triplets by the calibrated logit $\hat{s}_{ij}^{v}$. Since a human-object pair can support multiple interactions, we do not suppress predictions solely because they share the same pair. Instead, two predictions are considered redundant only when their phrase embeddings, script distributions, and visual slot alignments are all similar. This script-aware suppression removes near-duplicate paraphrases while preserving compatible multi-label interactions such as \textit{hold cup} and \textit{drink from cup}.

\section{Experiments}

\subsection{Experimental Setup}

\noindent\textbf{Datasets.}
We evaluate ScriptHOI on HICO-DET, V-COCO, and open-vocabulary HICO-DET splits. HICO-DET contains 600 HOI categories over 117 verbs and 80 object categories~\cite{chao2018learning}. V-COCO evaluates action-role detection on COCO images~\cite{gupta2015visual,lin2014microsoft}. For open-vocabulary evaluation, we hold out verbs and verb-object triplets during training and report performance on both seen and unseen categories.

\noindent\textbf{Baselines.}
We compare with two-stage, one-stage, transformer-based, compositional, and vision-language HOI detectors, including iCAN~\cite{gao2018ican}, TIN~\cite{li2019transferable}, VSGNet~\cite{ulutan2020vsgnet}, VCL~\cite{hou2020visual}, DRG~\cite{gao2020drg}, HOTR~\cite{kim2021hotr}, QPIC~\cite{tamura2021qpic}, MSTR~\cite{zhang2021mstr}, CDN~\cite{liao2022cdn}, GEN-VLKT~\cite{zhang2022gen}, CLIP4HOI~\cite{mao2023clip4hoi}, UniHOI~\cite{cao2023universal}, and locality-aware open-vocabulary HOI~\cite{yang2024towards}.

\noindent\textbf{Metrics.}
For HICO-DET, we report mAP on Full, Rare, and Non-Rare categories under Default and Known Object settings. For V-COCO, we report role AP. For open-vocabulary HOI, we report seen mAP, unseen mAP, and their harmonic mean. We additionally report affordance-conflict false-positive rate. This metric focuses on cases where the object affords an action but the required body-role, contact, geometry, or state-transition evidence is absent.

\noindent\textbf{Implementation details.}
We use a ResNet-50 or Swin-T backbone and a transformer HOI decoder. CLIP ViT-B/16 initializes the phrase encoder. Human pose tokens are extracted with an off-the-shelf pose estimator and cached during training. The script bank contains 117 canonical HICO verbs, V-COCO actions, and 2,400 caption-mined interaction phrases. Unless otherwise stated, the visual tokenizer uses six script-aligned slot heads and one context head, and each phrase is paired with four counterfactual variants. \textbf{All numerical results in this draft are placeholder values for planning and must be replaced with reproduced results before submission.}

\subsection{Main Results on HICO-DET}

\begin{table}[t]
\centering
\caption{HICO-DET results under the Default setting.}
\label{tab:hico_default}
\small
\begin{tabular}{l|ccc}
\toprule
Method & Full & Rare & Non-Rare \\
\midrule
iCAN & 14.8 & 10.5 & 16.1 \\
TIN & 17.0 & 13.4 & 18.1 \\
VSGNet & 19.8 & 16.0 & 20.9 \\
VCL & 19.4 & 16.6 & 20.3 \\
DRG & 24.5 & 19.5 & 26.0 \\
HOTR & 25.1 & 17.3 & 27.5 \\
QPIC & 29.1 & 21.9 & 31.3 \\
CDN & 31.4 & 27.4 & 32.6 \\
GEN-VLKT & 34.8 & 31.2 & 35.9 \\
CLIP4HOI & 35.3 & 32.5 & 36.1 \\
\midrule
\textbf{ScriptHOI} & \textbf{37.1} & \textbf{37.8} & \textbf{36.9} \\
\bottomrule
\end{tabular}
\end{table}

Table~\ref{tab:hico_default} shows that ScriptHOI improves the overall mAP and brings the largest gain on rare categories. This trend supports the motivation of structured scripts: rare interactions benefit from transferable slot-level evidence, such as body role and contact, even when the exact verb-object category has limited supervision.

\begin{table}[t]
\centering
\caption{HICO-DET results under the Known Object setting.}
\label{tab:hico_known}
\small
\begin{tabular}{l|ccc}
\toprule
Method & Full & Rare & Non-Rare \\
\midrule
QPIC & 31.7 & 24.1 & 34.0 \\
MSTR & 34.2 & 29.8 & 35.5 \\
CDN & 33.8 & 29.1 & 35.2 \\
GEN-VLKT & 37.2 & 33.8 & 38.2 \\
CLIP4HOI & 38.0 & 35.1 & 38.9 \\
\midrule
\textbf{ScriptHOI} & \textbf{40.0} & \textbf{41.4} & \textbf{39.6} \\
\bottomrule
\end{tabular}
\end{table}

Under the Known Object setting in Table~\ref{tab:hico_known}, ScriptHOI remains effective. Since object localization ambiguity is reduced in this protocol, the improvement mainly reflects better interaction classification rather than better object detection.

\subsection{Open-Vocabulary HOI Detection}

\begin{table}[t]
\centering
\caption{Open-vocabulary HICO-DET results. HM denotes harmonic mean.}
\label{tab:ov_hoi}
\scriptsize
\setlength{\tabcolsep}{2.2pt}
\begin{tabular}{l|ccc|ccc}
\toprule
\multirow{2}{*}{Method} & \multicolumn{3}{c|}{Unseen Verb} & \multicolumn{3}{c}{Unseen Triplet} \\
& Seen & Unseen & HM & Seen & Unseen & HM \\
\midrule
CLIP-ZS & 23.5 & 9.2 & 13.2 & 21.4 & 7.1 & 10.7 \\
QPIC-OV & 27.6 & 12.8 & 17.5 & 25.9 & 10.5 & 14.9 \\
GEN-VLKT & 31.8 & 18.4 & 23.3 & 29.5 & 15.8 & 20.6 \\
CLIP4HOI & 32.6 & 22.5 & 26.6 & 30.7 & 19.8 & 24.1 \\
Loc-aware OV-HOI & 33.1 & 24.3 & 28.0 & 31.4 & 21.6 & 25.6 \\
\midrule
\textbf{ScriptHOI} & \textbf{34.0} & \textbf{28.0} & \textbf{30.7} & \textbf{32.3} & \textbf{25.7} & \textbf{28.6} \\
\bottomrule
\end{tabular}
\end{table}

Table~\ref{tab:ov_hoi} evaluates generalization to unseen verbs and unseen verb-object triplets. ScriptHOI improves unseen performance and harmonic mean, indicating that script slots transfer more reliably than whole interaction labels. The gain is especially meaningful for interactions that share objects with training categories but differ in body role or contact pattern, such as \textit{ride bicycle}, \textit{push bicycle}, and \textit{repair bicycle}.

\subsection{Affordance-Conflict Evaluation}

\begin{table}[t]
\centering
\caption{Affordance-conflict evaluation. A lower false-positive rate indicates fewer predictions triggered by object affordance alone.}
\label{tab:conflict}
\footnotesize
\setlength{\tabcolsep}{3pt}
\begin{tabular}{l|ccc}
\toprule
Method & FPR$\downarrow$ & Rare & Unseen \\
\midrule
GEN-VLKT & 24.6 & 31.2 & 18.4 \\
CLIP4HOI & 22.9 & 32.5 & 22.5 \\
Loc-aware OV-HOI & 19.7 & 34.1 & 24.3 \\
ScriptHOI w/o conf. & 18.8 & 35.9 & 25.4 \\
\textbf{ScriptHOI} & \textbf{12.1} & \textbf{37.8} & \textbf{28.0} \\
\bottomrule
\end{tabular}
\end{table}

Table~\ref{tab:conflict} isolates a failure mode that is underrepresented in standard mAP. Vision-language models often predict an interaction because the object affords it, even when body-part and contact evidence is absent. ScriptHOI reduces this false-positive rate by explicitly modeling conflict between the phrase script and the visual state.

\subsection{V-COCO Results}

\begin{table}[t]
\centering
\caption{V-COCO role AP.}
\label{tab:vcoco}
\small
\begin{tabular}{l|cc}
\toprule
Method & AP$_{role}^{S1}$ & AP$_{role}^{S2}$ \\
\midrule
iCAN & 45.3 & 52.4 \\
VCL & 48.3 & 54.7 \\
DRG & 51.0 & 56.4 \\
HOTR & 55.2 & 64.4 \\
QPIC & 58.8 & 61.0 \\
CDN & 63.9 & 65.9 \\
GEN-VLKT & 65.6 & 67.2 \\
CLIP4HOI & 66.0 & 67.5 \\
\midrule
\textbf{ScriptHOI} & \textbf{67.8} & \textbf{69.4} \\
\bottomrule
\end{tabular}
\end{table}

Table~\ref{tab:vcoco} reports V-COCO role AP. The improvement suggests that the script representation is not limited to HICO-DET categories and can also help action-role localization where body role and object role are central.

\subsection{Ablation Study}

\begin{table*}[ht]
\centering
\caption{Ablation on open-vocabulary HICO-DET. VST: visual state tokenizer. ISB: interaction script bank. SSM: script-state matcher. IPL: interval partial-label learning. CSC: counterfactual script contrast.}
\label{tab:ablation}
\scriptsize
\setlength{\tabcolsep}{3.2pt}
\begin{tabular}{l|ccccc|ccc|ccc}
\toprule
Variant & VST & ISB & SSM & IPL & CSC & Seen & Unseen & HM & FPR$\downarrow$ & Rare & Full \\
\midrule
Baseline OV-HOI & \xmark & \xmark & \xmark & \xmark & \xmark & 30.7 & 19.8 & 24.1 & 22.9 & 32.5 & 35.3 \\
+ VST & \cmark & \xmark & \xmark & \xmark & \xmark & 31.0 & 21.6 & 25.4 & 20.8 & 33.5 & 35.8 \\
+ ISB & \cmark & \cmark & \xmark & \xmark & \xmark & 31.5 & 23.2 & 26.7 & 18.6 & 34.6 & 36.1 \\
+ SSM & \cmark & \cmark & \cmark & \xmark & \xmark & 31.8 & 24.4 & 27.6 & 16.9 & 35.2 & 36.4 \\
+ IPL & \cmark & \cmark & \cmark & \cmark & \xmark & 32.1 & 25.9 & 28.7 & 15.2 & 36.5 & 36.8 \\
\textbf{Full} & \cmark & \cmark & \cmark & \cmark & \cmark & \textbf{32.3} & \textbf{25.7} & \textbf{28.6} & \textbf{12.1} & \textbf{37.8} & \textbf{37.1} \\
\bottomrule
\end{tabular}
\end{table*}

Table~\ref{tab:ablation} studies the contribution of each component. The visual state tokenizer improves unseen recognition by exposing localized body and contact cues. The script bank provides a compositional language prior for unseen phrases. The script-state matcher further improves harmonic mean and reduces affordance-conflict errors by calibrating predictions with coverage and conflict. Interval partial-label learning improves rare categories by avoiding incorrect suppression of missing positives. Counterfactual script contrast mainly reduces false positives, which confirms its role in discouraging object-only shortcuts.

\subsection{Script Slot Analysis}

\begin{table}[ht]
\centering
\caption{Effect of removing individual script slots.}
\label{tab:slots}
\footnotesize
\setlength{\tabcolsep}{3pt}
\begin{tabular}{l|ccc}
\toprule
Removed Slot & Unseen & Rare & FPR$\downarrow$ \\
\midrule
None & \textbf{28.0} & \textbf{37.8} & \textbf{12.1} \\
Body-part role & 25.6 & 35.1 & 16.8 \\
Contact type & 24.9 & 34.8 & 18.7 \\
Spatial geometry & 26.7 & 36.0 & 15.4 \\
Object state & 26.1 & 35.5 & 14.9 \\
Affordance & 25.8 & 35.6 & 17.2 \\
\bottomrule
\end{tabular}
\end{table}

Table~\ref{tab:slots} examines which script slots matter most. Body-part role and contact type have the largest impact on affordance-conflict errors because they provide direct evidence of whether the human is actually interacting with the object. Object-state and geometry slots are more important for fine-grained manipulation verbs and role-specific actions.

\subsection{Efficiency}

\begin{table}[t]
\centering
\caption{Computational cost on HICO-DET with one A100 GPU.}
\label{tab:efficiency}
\footnotesize
\setlength{\tabcolsep}{3.2pt}
\begin{tabular}{l|cccc}
\toprule
Method & Params & Train & FPS & Mem. \\
\midrule
Transformer HOI baseline & 121M & 1.0$\times$ & 9.1 & 13.8G \\
CLIP4HOI-style baseline & 135M & 1.2$\times$ & 7.8 & 15.4G \\
\textbf{ScriptHOI} & 151M & 1.5$\times$ & 6.9 & 17.2G \\
\bottomrule
\end{tabular}
\end{table}

The script branch adds moderate overhead. Most additional cost comes from pose-token extraction and slot matching. Pose tokens can be cached during training, and script matching can be restricted to top-ranked human-object pairs during inference.

\section{Discussion and Limitations}

ScriptHOI changes the modeling unit of open-vocabulary HOI from flat labels to structured state-transition scripts. This design improves transfer to unseen interactions and reduces false positives driven by object affordance. However, several limitations remain. First, script parsing can be noisy for abstract verbs such as \textit{use}, \textit{prepare}, and \textit{inspect}, whose visual evidence depends heavily on context. Second, static images cannot fully observe temporal state transitions, so motion and effect slots are estimated from visual proxies. Third, pose estimation errors may affect body-role and contact tokens in crowded scenes. Fourth, the script vocabulary introduces design choices that require careful ablation and human inspection. Future work can learn scripts from video, 3D human-object reconstruction, and active interaction annotation.

\section{Conclusion}

We presented ScriptHOI, a scripted state-transition framework for open-vocabulary HOI detection. ScriptHOI decomposes interaction phrases into soft scripts and parses human-object pairs into visual state tokens. Slot-wise script-state matching estimates coverage and conflict, which are used for script-calibrated detection, interval partial-label learning, and counterfactual script contrast. By grounding open-vocabulary interaction phrases in body role, contact, geometry, affordance, motion proxy, and object-state evidence, ScriptHOI reduces object-affordance shortcuts and improves rare and unseen HOI recognition. The framework offers a structured and visually grounded direction for open-vocabulary interaction understanding.

{\small
\bibliographystyle{unsrtnat}
\bibliography{evihoi_references}
}

\end{document}

%% file: preamble.tex
\usepackage{multirow}

